\documentclass{article}
\usepackage{ijcai22}

\usepackage{times}
\usepackage{soul}
\usepackage{url}
\usepackage[hidelinks]{hyperref}
\usepackage[utf8]{inputenc}
\usepackage[small]{caption}
\usepackage{graphicx}
\usepackage{amsmath}
\usepackage{amsthm}
\usepackage{booktabs}
\usepackage{algorithm}
\usepackage{algorithmic}
\usepackage{amsmath}
\usepackage{multirow,bm}
\usepackage{amssymb}
\usepackage{amsthm}
\usepackage{bbm}
\usepackage{stmaryrd}
\usepackage{booktabs}
\usepackage{wrapfig}
\usepackage{pdfpages}

\newtheorem{theorem}{Theorem}
\newtheorem{innercustomthm}{Theorem}
\newenvironment{customthm}[1]
  {\renewcommand\theinnercustomthm{#1}\innercustomthm}
  {\endinnercustomthm}
  
\newtheorem{innercustomdef}{Definition}
\newenvironment{customdef}[1]
  {\renewcommand\theinnercustomdef{#1}\innercustomdef}
  {\endinnercustomdef}

\newenvironment{sproof}{%
  \proof}{\endproof}

\newcommand{\domx}{\mathcal{X}}
\newcommand{\domy}{\mathcal{Y}}

\newcommand{\vecx}{\boldsymbol{x}}
\newcommand{\vecy}{\boldsymbol{y}}

\newcommand{\vech}{\boldsymbol{h}}

\title{None Class Ranking Loss for Document-Level Relation Extraction}

\author{
Yang Zhou$^1$
\and
Wee Sun Lee$^1$
\affiliations
$^1$School of Computing, National University of Singapore
\emails
\{zhouy, leews\}@comp.nus.edu.sg
}

\begin{document}

\maketitle

\begin{abstract}
    Document-level relation extraction (RE) aims at extracting relations among entities expressed across multiple sentences, which can be viewed as a multi-label classification problem. 
    In a typical document, most entity pairs do not express any pre-defined relation and are labeled as ``none" or ``no relation".
    For good document-level RE performance, it is crucial to distinguish such \textit{none} class instances (entity pairs) from those of pre-defined classes (relations).
    However, most existing methods only estimate the probability of pre-defined relations independently without considering the probability of ``no relation". 
    This ignores the context of entity pairs and the label correlations between the none class and pre-defined classes, leading to sub-optimal predictions.
    To address this problem, we propose a new multi-label loss that encourages large \textit{margins} of label confidence scores between each pre-defined class and the none class, which enables captured label correlations and context-dependent thresholding for label prediction. 
    To gain further robustness against positive-negative imbalance and mislabeled data that could appear in real-world RE datasets, we propose a margin regularization and a margin shifting technique.
    Experimental results demonstrate that our method significantly outperforms existing multi-label losses for document-level RE and works well in other multi-label tasks such as emotion classification when none class instances are available for training.
\end{abstract}

\section{Introduction} \label{sec:intro}
Identifying relationships among entities from unstructured text, namely relation extraction (RE), is a fundamental task in information extraction. Previous works have studied sentence-level RE that aims at extracting relations between two entities in a single sentence \cite{Miwa2016EndtoEndRE,Zhang2018GraphCO}. More recently, document-level RE, which aims to identify the relations of various entity pairs expressed across \textit{multiple sentences}, has received increasing research attention \cite{Yao2019DocREDAL,yu2020dialogue}. 

Since one entity pair can express multiple relations in a document, document-level RE can be formulated as a \textit{multi-label} classification problem. Different from conventional multi-label settings where each instance at least has one positive pre-defined label, most instances (entity pairs) in document-level RE express \textit{none} of the pre-defined relations, where all the pre-defined labels are \textit{known to be negative}. For good document-level RE performance, it is crucial to distinguish such \textit{none} class instances (entity pairs) from those of pre-defined classes (relations).

However, most existing methods adopt binary cross entropy (BCE) loss for document-level RE. They only estimate the probability of pre-defined relations \textit{independently} without considering the probability of ``no relation". This ignores the fact that the probabilities of both the \textit{none} class and pre-defined classes are important for multi-label prediction, as they are highly correlated (the increased probability of the none class implies the decreased probabilities of all the pre-defined classes). Without knowing the none class probability, whether an entity pair has certain relations needs to be determined via global thresholding (i.e., selecting labels whose confidence scores are higher than a single threshold as final predictions). This fails to take the context of each entity pair into account and may produce either too many or too few labels, leading to sub-optimal predictions.

In this work, we propose a new multi-label performance measure and its surrogate loss, called None Class Ranking Loss (NCRL), to address the aforementioned disadvantages of BCE in document-level relation extraction. NCRL aims to maximize the \textit{margin} of label confidence scores between the none class and each pre-defined class, such that positive labels are ranked above \textit{the none class label} and negative labels are ranked below. In this way, the none class score becomes an instance-dependent threshold that treats labels being ranked above the none class as positive, and the label margin reflects the uncertainty of such prediction. By formulating the probability of each class as label margins, NCRL not only takes the context of entity pairs in multi-label prediction but also (as will be shown later) enables captured label correlations without directly comparing the scores of positive and negative labels. 
Moreover, we theoretically justify the effectiveness of NCRL by proving that NCRL satisfies \textit{Bayes consistency} w.r.t. the targeted multi-label performance measure, which is a desirable property for any surrogate loss in multi-label learning.

In a document, entity pairs generally have a few positive relations and many negative ones. This positive-negative imbalance makes multi-label losses focus more on negative samples and down-weight the contributions from the rare positive samples \cite{ben2020asymmetric}. 
Another practical problem of document-level RE is that entity pairs are prone to being mislabeled as ``no relation", which results in deteriorated classification performance.
To alleviate the positive-negative imbalance problem, we propose a margin regularization technique, which penalizes negative samples whose confidence scores are far below that of the none class. For robustness against mislabeled samples, we propose a margin shifting approach to attenuate the impact of negative samples with a very large score, which are suspected as mislabeled. This helps to reduce the label noise in document-level RE datasets. 

Our NCRL loss is general enough and can be applied to other multi-label tasks when none class instances are available for training. Experimental results on two document-level RE datasets, DocRED \cite{Yao2019DocREDAL} and DialogRE \cite{yu2020dialogue}, and one fine-grained emotion classification dataset, GoEmotions \cite{demszky-etal-2020-goemotions}, show that NCRL achieves significantly better results than existing multi-label losses. The main contributions of this paper are summarized as follows:
\begin{itemize}
    \item We propose a new multi-label performance measure along with its surrogate loss, NCRL, that prefers large label margins between the none class and each pre-defined class. Our loss captures the label correlations and enables context-dependent thresholds for multi-label prediction.
    \item We prove that NCRL satisfies multi-label consistency, which guarantees that minimizing NCRL can achieve the expected objective.
    \item We propose a margin regularization and a margin shifting technique to alleviate positive-negative sample imbalance and gain further robustness against mislabeled samples, respectively.
\end{itemize}

\section{Related Work}

\textbf{Multi-label learning.}
Multi-label classification is a well-studied problem with rich literature. Here, we focus on loss functions for multi-label learning.
BCE is the most popular multi-label loss, which reduces the multi-label problem into a number of independent binary (one-vs-all) classification tasks, one for each label. 
Another common multi-label loss function is pairwise ranking loss, which transforms multi-label learning into a ranking problem via pairwise (one-vs-one) comparison.
Many attempts have been made to improve these two types of multi-label loss functions \cite{furnkranz2008multilabel,yeh2017learning,li2017improving,durand2019learning,xu2019robust,wu-etal-2019-learning,ben2020asymmetric}. However, as far as we know, none of the existing losses as well as theoretical studies consider none class instances, which commonly appear in real-world multi-label applications.

\textbf{Document-level relation extraction.}
Existing works on document-level RE mainly focus on learning better entity pair representations to improve classification performance. Some representative approaches include incorporating long-distance entity interactions into recurrent neural networks or transformer-based models \cite{christopoulou2019connecting,nan-etal-2020-reasoning}, aggregating local and global entity information with graph neural networks \cite{wang2020global,zeng2020double,docre_aaai21,dai2021coarse}, and integrating latent word semantics with pre-trained language models \cite{zhou2021atlop,xu2021entity}.

All the aforementioned document-level RE methods, except \cite{zhou2021atlop}, use BCE as the objective function for training and do not consider the none class. 
ATL \cite{zhou2021atlop} is probably the most related method to our NCRL loss function. Both ATL and NCRL treat the none class label as an adaptive threshold to separate positive relations from negative ones, leading to context-dependent multi-label prediction. However, the proposed NCRL can learn a more discriminative classifier with theoretical guarantees and consistently outperforms ATL in our experiments. For clarity, we defer the detailed comparisons between NCRL and ATL in Sec. \ref{sec:comp loss}.

\section{None Class Ranking Loss}
In this section, we first introduce None Class Ranking Error (NCRE), a new multi-label performance measure that takes into account the none class in multi-label learning. Then, we propose None Class Ranking Loss (NCRL), which is a surrogate loss of NCRE with Bayes consistency. Finally, we propose a margin regularization and a margin shifting strategy to address the positive-negative imbalance and mislabeling problems that commonly appear in real-world RE datasets.

\subsection{Problem Statement}

Document-level RE is essentially a multi-label classification problem, where each entity pair is an instance and the associated relations are label samples. Let $\domx$ be an instance space and $\domy=\{0, 1\}^{K}$ be a label space, where $K$ is the number of pre-defined classes. An instance $\vecx \in \domx$ is associated with a subset of labels, identified by a binary vector $\vecy \in \domy = (y_1, \ldots, y_K)$, where $y_i = 1$ if the $i$-th label is positive for $\vecx$, and $y_i=0$ otherwise. In particular, none class instances can be represented by setting all the labels to be negative ($y_i=0, i=1,\ldots,K$). To simplify the notation and facilitate deriving the proposed NCRL, we introduce the none class and assign it with the $0$-th label $y_0$ without loss of generality. Then none class instances can be simply labeled with $y_0=1$. 

\subsection{Performance Measure}
Different from document-level RE, conventional multi-label datasets often assume that each instance at least has one positive label, which means that none class instances are not available for training. For this reason, existing multi-label performance measures only focus on the predictions of pre-defined labels individually, and do not consider how well the none class is predicted. Nevertheless, the probabilities of the pre-defined classes are highly correlated with the probability of the none class. 
Failing to consider such label correlations in the performance measure could overlook bad predictions that should have been penalized during training, and thus increases the risk of misclassification, especially for document-level RE, where most entity pairs are from the none class.

To better measure the performance of document-level RE, we transform the multi-label prediction of each instance into a label ranking, where positive pre-defined labels should be ranked higher than the none class label, while negative ones should be ranked below. Formally, we propose NCRE, a new multi-label performance measure as follows:
\begin{align} \label{NCRL}
    \ell_\textsc{NA}(\vecy,\bm{f})=&\sum_{i=1}^{K}  \Big( \llbracket y_i > 0 \rrbracket \llbracket f_i < f_0 \rrbracket \nonumber \\
    &+ \llbracket y_i \leq 0 \rrbracket \llbracket f_i > f_0 \rrbracket + \frac{1}{2} \llbracket f_i = f_0 \rrbracket \Big),
\end{align}
where $f_i$ is the confidence score of the $i$-th class, and $\llbracket \pi \rrbracket$ is the mapping that returns $1$ if predicate $\pi$ is true and $0$ otherwise. 
NCRE treats the none class label as an adaptive threshold that separates positive labels from negative ones, and counts the number of reversely ordered label pairs between the none class and each pre-defined class, which considers errors in ranking the none class and pre-defined labels simultaneously.

\subsection{NCRL Surrogate Loss}

Although our goal is to find an optimal classifier w.r.t. NCRE, it is difficult to minimize NCRE directly. For practical multi-label learning, we need to minimize a surrogate loss of NCRE instead, which should be simple and easy for optimization. To this end, we propose None Class Ranking Loss (NCRL) as follows:
\begin{equation}
    \label{NCRL surrogate}
    L_\textsc{NA}(\vecy,\bm{f}) = - \sum_{i=1}^{K} \Big( y_i \cdot \log \sigma(m_i^{+}) + (1 - y_i) \cdot \log \sigma(m_i^{-}) \Big),
\end{equation}
where $m_i^{+} = f_i - f_0$ and $m_i^{-} = f_0 - f_i$ represent \textit{positive and negative label margins}, respectively, and $\sigma(x)$ is the sigmoid function that transforms label margins into probability outputs within $[0, 1]$. Intuitively, minimizing NCRL encourages positive (negative) labels to have higher (lower) scores than that of the none class as much as possible. At the test time, we can return labels with higher scores than $f_0$ as predicted labels or return \textsc{NA} if such labels do not exist.

\textbf{Captured correlations.} 
BCE assumes pre-defined classes are independent and thus ignores label correlations and dependencies. Pairwise ranking loss is commonly used to capture label correlations in multi-label learning, which aims to maximize the margin between each pair of positive and negative labels. Compared with BCE and pairwise ranking loss, NCRL relaxes the independent assumption of BCE and captures label correlations by connecting the probability of each pre-defined class $\mathbb{P}(y_i=1|\bm{x})$ to the label margins $m_i^{+} = f_i - f_0$. In this way, maximizing $\mathbb{P}(y_i=1|\bm{x})$ not only leads to increased $f_i$ but also decreased $f_0$, which in turn affects the margins (probabilities) of all the other classes. 

\begin{figure}[t!]
\centering
\includegraphics[width=0.8\columnwidth]{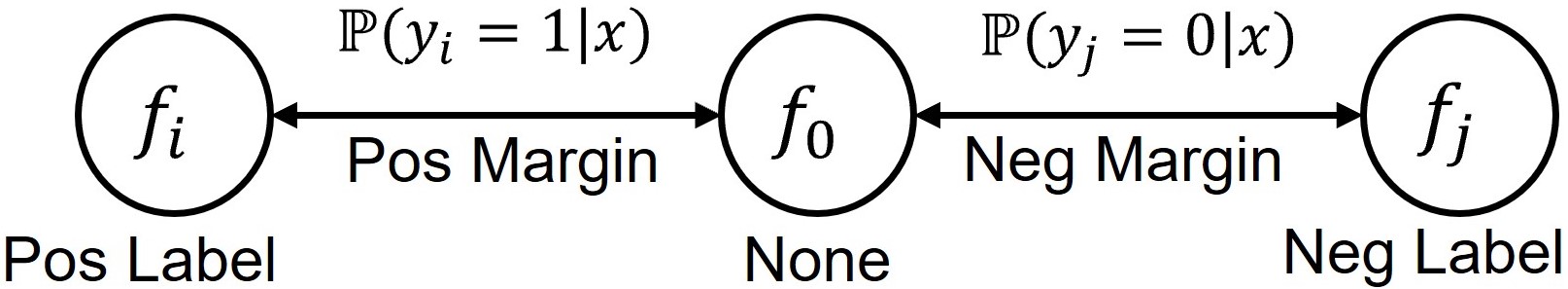} 
\caption{\small An illustration of NCRL. For $y_i=1$, NCRL maximizes the positive margin $f_i-f_0 \propto \mathbb{P}(y_i=1|\bm{x})$ (see Theorem \ref{thm:consist}). For $y_j=0$, NCRL maximizes the negative margin $f_0-f_j \propto p(y_j=0|\bm{x})$. In this way, the margin between positive and negative labels $f_i - f_j = (f_i - f_0) + (f_0 - f_j)$ is also maximized.}
\label{NCRL_Ill}
\end{figure}

Moreover, thanks to the introduced none class, the margin between any pair of pre-defined labels can be induced from the sum of $m_i^{+}$ and $m_j^{-}$, i.e., $f_i - f_j = (f_i - f_0) + (f_0 - f_j)$. Therefore, NCRL captures pairwise label correlations and \textit{indirectly} maximizes the margin between every pair of positive and negative labels without expensive pairwise comparisons. Figure \ref{NCRL_Ill} illustrates how NCRL learns a proper label ranking by separating positive and negative labels via $f_0$.

\textbf{Consistency analysis.}
Now we theoretically justify the effectiveness of NCRL by proving that NCRL (\ref{NCRL surrogate}) is \textit{Bayes consistent} w.r.t. NCRE (\ref{NCRL}). Bayes consistency is an important property for a surrogate loss. It guarantees that the classifier obtained by minimizing the surrogate loss converges to the one with minimal multi-label risk, and thus is likely to achieve good classification performance w.r.t. the corresponding multi-label performance measure.

Given an instance $\vecx$, let $\Delta_i = \mathbb{P}(y_i = 1|\vecx)$ be the marginal probability when the $i$-th label is positive. Then the conditional risk w.r.t. NCRE (\ref{NCRL}) can be written as follows:
\begin{align} \label{NCRL risk}
    R_{{\ell}_\textsc{na}}(\mathbb{P}, \bm{f}) =& \mathbb{E}[ \ell_{\textsc{na}}(\mathbb{P}, \bm{f}) |\vecx] =  \sum_{i=1}^{K}  \Big( \Delta_i \llbracket f_i < f_0 \rrbracket \nonumber \\ &+ (1 - \Delta_i) \llbracket f_i > f_0 \rrbracket + \frac{1}{2} \llbracket f_i = f_0 \rrbracket \Big),
\end{align}
which corresponds to the \textit{expected penalty} for an instance $\vecx$ when $\bm{f}(\vecx)$ is used as the score function. From (\ref{NCRL risk}), the Bayes optimal score function $\bm{f}_{\ell_\textsc{na}}^*$ that minimizes the multi-label risk (\ref{NCRL risk}) is given by:
\begin{equation} \label{Bayes opt}
    \bm{f}_{\ell_\textsc{na}}^* \in \{ \bm{f}:  f_i > f_0 \text{ if }  \Delta_i > \frac{1}{2}, \text{ and } f_i < f_0 \text{ if } \Delta_i < \frac{1}{2} \}.
\end{equation}

\begin{theorem}
    \label{thm:consist}
    NCRL (\ref{NCRL surrogate}) is Bayes consistent w.r.t. NCRE (\ref{NCRL}).
\end{theorem}
\begin{sproof}
    We prove the Bayes consistency of NCRL by showing that given an instance, the optimal label margin $f_i^* - f_0^*$ obtained by minimizing NCRL \textit{coincides with} the marginal label probability $\Delta_i = \mathbb{P}(y_i = 1|\vecx)$, such that $f_i^* - f_0^* > 0 \Rightarrow \Delta_i > \frac{1}{2}$, which is an optimal solution with the minimal NCRE risk according to (\ref{Bayes opt}). The complete proof is provided in Appendix.
\end{sproof}

\subsection{Margin Regularization and Shifting}
\textbf{Margin regularization.} 
In real-world document-level RE datasets, it is common that only a small number of entity pairs express pre-defined relations, implying that the number of positive samples per relation is much larger than that of negative ones. Such positive-negative sample imbalance makes NCRL push a large number of negative samples away from the threshold $f_0$ to achieve a low training loss. As a result, training with imbalanced data tends to make negative predictions for pre-defined labels with rare positive samples.

To address this problem, we propose to control the average margin and design the following regularization term:
\begin{equation} \label{none reg}
    L_{\textsc{NA}_0}(\vecy,\bm{f}) = - y_0 \cdot \log \sigma(m_0^+) - (1 - y_0) \cdot \log \sigma(m_0^-) 
\end{equation}
where $m_0^+ = f_0 - \frac{1}{K} \sum_{i=1}^K f_i$ and $m_0^- = \frac{1}{K} \sum_{i=1}^K f_i - f_0$ are the average positive and negative margins, respectively. For instances with $y_0 = 0$, the margin regularization $L_{\textsc{NA}_0}(\vecy,\bm{f})$ penalizes the average score over all pre-defined labels if it is smaller than $f_0$, and thus avoids over-suppressing scores of negative samples. On the other hand, for none class instances with $y_0 = 1$, the threshold $f_0$ should be larger than the largest score among all the pre-defined labels. However, directly maximizing the margin $f_0 - \max_{i\neq0} f_i$ can lead to unstable results, as the pre-defined label with $\max_{i\neq0} f_i$ would keep changing during training. Instead, $L_{\textsc{NA}_0}(\vecy,\bm{f})$ maximizes the average margin $f_0 - \frac{1}{K} \sum_{i=1}^K f_i$, which is an upper-bound of $f_0 - \max_{i\neq0} f_i$, to encourage $f_0$ being at the top of the label ranking.

\textbf{Margin shifting.} Another practical issue in document-level RE is the mislabeling problem. As a document can describe various contents, it is very difficult for a human annotator to annotate all relations accurately and extensively for each entity pair. As a result, real-world document-level RE datasets often contain considerable noise in labeling. Among mislabeled data, false negative samples are particularly common in practice. 

To gain more robustness against mislabeled samples, we propose to shift the transformed negative margin $\sigma(m^-_i)$ in NCRL (\ref{NCRL surrogate}) as follows:
\begin{equation} \label{shifting}
    p_i^- = \max(\sigma(m^-_i) + \gamma, 1), i = 0, 1, \ldots, K,
\end{equation}
where $0 < \gamma < 1 $ is a hyper-parameter. Such margin shifting approach increases the probability of all negative samples by a factor $\gamma$ such that the impact of very hard negative samples with small $\sigma(m^-_i) = \mathbb{P}(y_i=0|\vecx)$, which are suspected as mislabeled, are attenuated during training. In addition, very easy samples with $\sigma(m^-_i) > 1 - \gamma$ get a zero loss and are ignored, which makes NCRL focus more on harder samples.

Although similar shifting approaches have been proposed \cite{lin2017focal,ben2020asymmetric,wu2020distribution}, they do not consider the none class label and shift the \textit{score} $f_i$ rather than the \textit{margin} $m_i^-$. Since the none class score $f_0$ \textit{aligns} the margins $m_i^-$ of pre-defined labels in the same range, margin shifting is more suitable than score shifting in adjusting the prediction probability for different training samples. Armed with the margin regularization and shifting, our final loss is given as follows:
\begin{equation}
    \label{final NCRL}
    L_\textsc{NA}(\vecy,\bm{f}) = - \sum_{i=0}^{K} \Big( y_i \cdot \log \sigma(m_i^{+}) + (1 - y_i) \cdot \log p_i^- \Big).
\end{equation}

\subsection{Comparison with ATL} \label{sec:comp loss}

There are a few related works that consider the none class in document-level RE. To the best of our knowledge, adaptive thresholding loss (ATL) \cite{zhou2021atlop} is the most related method to our NCRL. 
ATL also treats the none label class as a learnable threshold and encourages positive labels to be ranked above the none class label and negative labels to be ranked below for document-level RE. Specifically, ATL is given as follows:
\begin{align*}
    L_{ATL}(\vecy,\bm{f}) =& -\sum_{i: y_i = 1} \log \left( \frac{\exp\left(f_i\right)}{\exp(f_0) + \sum_{j: y_j=1} \exp \left(f_{j}\right)} \right) \\ & - \log \left( \frac{\exp\left(f_0\right)}{\exp(f_0) + \sum_{j: y_j=0} \exp \left(f_{j}\right)} \right), 
\end{align*}
where $i,j = 1,\ldots,K$. 

Although ATL and NCRL share the same spirit of learning the none class label, they essentially learn distinct label rankings. ATL is constructed in two terms. The first term prefers the none class score $f_0$ being \textit{ranked at the bottom} among $f_0$ and the scores of positive pre-defined labels, and the second term prefers $f_0$ being \textit{ranked at the top} among $f_0$ and the scores of negative pre-defined labels. ATL only focuses on positioning $f_0$ between positive and negative sets of labels, whereas fails to maximize the margin between each pair of positive and negative labels, which is desirable for obtaining discriminative classifiers. Due to the softmax function in the first term of ATL, increasing the score of one positive labels $f_i$ will penalize the scores of all the other positive labels, and could leads to large loss values, which is not reasonable in the multi-label setting. On the contrary, our NCRL separates positive labels from negative ones as much as possible by maximizing the label margins, and enjoys the theoretical guarantee of Bayes consistency. In addition, NCRL is equipped with the margin regularization and shifting to alleviate the positive-negative imbalance and mislabeling problems, which are not applicable for ATL.

\section{Experiments}
In this section, we evaluate NCRL on two document-level RE datasets. To demonstrate the effectiveness of NCRL for other multi-label tasks, we also test NCRL for fine-grained emotion classification.

\subsection{Experimental Setups} \label{subsec:setup}
\textbf{Datasets.}
\textit{DocRED}~\cite{Yao2019DocREDAL} is a large-scale document-level RE dataset, which is constructed from Wikipedia articles. DocRED consists of 5053 documents with 96 pre-defined relations, where 3053 documents are used for training, 1000 for development, and 1000 for testing. DocRED also provides a large distantly supervised dataset, which includes 101,873 documents with noisy labels. \textit{DialogRE}~\cite{yu2020dialogue} is a dialogue-based RE dataset, which is constructed from the transcripts of the American TV situation comedy, Friends. DialogRE consists of 1788 dialogues with 36 pre-defined relations, where 60\% of dialogues are used for training, 20\% for development, and 20\% for testing. \textit{GoEmotions}~\cite{demszky-etal-2020-goemotions} is an emotion classification dataset of 58009 Reddit comments labeled for 27 emotion categories or Neutral, where 80$\%$ data are used for training, 10$\%$ for development, and 10$\%$ for testing. 

\begin{table*}[!t]
\centering
\small
    \begin{tabular}{p{7.5cm}cccc}
        \toprule
        \textbf{Model} & \multicolumn{2}{c}{\textbf{Dev}} & \multicolumn{2}{c}{\textbf{Test}} \\
        & Ign $F_1$ & $F_1$ & Ign $F_1$ & $F_1$ \\
        \midrule
        HIN-BERT$_{\text{Base}}$~\cite{Tang2020HINHI}& 54.29& 56.31& 53.70& 55.60 \\
        CFER-BERT$_{\text{Base}}$~\cite{dai2021coarse} & 58.00 & 60.06 & 57.89 & 59.82 \\
        Coref-RoBERTa$_{\text{Large}}$~\cite{Ye2020CoreferentialRL}& 57.35& 59.43& 57.90& 60.25 \\
        SSAN-RoBERTa$_{\text{Large}}$~\cite{xu2021entity} & 60.25 & 62.08 & 59.47 & 61.42 \\
        \midrule
        BCE + ATLOP-RoBERTa$_{\text{Large}}$ & 61.02 $\pm$ 0.21  & 63.19 $\pm$ 0.20 & 61.58 & 63.46 \\  
        ATL + ATLOP-RoBERTa$_{\text{Large}}$~\cite{zhou2021atlop} & 61.32 $\pm$ 0.14& 63.18 $\pm$ 0.19& 61.39& 63.40 \\  
        NCRL + ATLOP-RoBERTa$_{\text{Large}}$ & 62.21 $\pm$ 0.22 & 64.18 $\pm$ 0.20 & 61.94 & 64.14\\
        \midrule
        BCE + ATLOP-DeBERTa$_{\text{Large}}$ & 61.92 $\pm$ 0.13  & 63.96 $\pm$ 0.15 & 61.83 & 63.92 \\  
        ATL + ATLOP-DeBERTa$_{\text{Large}}$ & 62.16 $\pm$ 0.15  & 64.01 $\pm$ 0.12 &  62.12 & 64.08 \\         
        NCRL + ATLOP-DeBERTa$_{\text{Large}}$ & \textbf{62.98 $\pm$ 0.18}& \textbf{64.79 $\pm$ 0.13}& \textbf{63.03} &  \textbf{64.96} \\
        \midrule
        \midrule
        With pre-training on the distantly supervised dataset & \multicolumn{4}{c}{} \\        
        BCE + ATLOP-DeBERTa$_{\text{Large}}$ & 65.05 $\pm$ 0.24  & 66.71 $\pm$ 0.22 & 64.91 & 66.69 \\  
        ATL + ATLOP-DeBERTa$_{\text{Large}}$ & 64.34 $\pm$ 0.12  & 66.18 $\pm$ 0.15 &  63.95 & 65.90 \\         
        NCRL + ATLOP-DeBERTa$_{\text{Large}}$ & \textbf{66.11 $\pm$ 0.14}& \textbf{67.92 $\pm$ 0.14}& \textbf{65.81} &  \textbf{67.53} \\
        \bottomrule
    \end{tabular}
    \caption{Classification results (\%) on the DocRED dataset.}
    \label{tab::docred_results}
    \vspace*{-10pt}
\end{table*}

\textbf{Implementation details.}
For each dataset, we adopt an encoding model upon pre-trained language models for representation learning. The classification results are then obtained by training the encoding model with different loss functions. We use Huggingface's Transformers~\cite{wolf2020transformers} to implement all the models. We use AdamW~\cite{Loshchilov2019DecoupledWD} as the optimizer with learning rates $\in \{1\mathrm{e}{-5}, 2\mathrm{e}{-5}, \ldots 5\mathrm{e}{-5}\}$, and apply a linear warmup~\cite{Goyal2017AccurateLM} at the first 10\% steps followed by a linear decay to 0. The number of training epochs is selected from $\{5, 8, 10, 20, 30\}$. At the test time, the best checkpoint on the development set is used for the final prediction. For BCE, we follow the same settings with ATL \cite{zhou2021atlop} to use a global threshold for multi-label prediction and select the threshold from $\{0.1, 0.2, ..., 0.9\}$. For NCRL, the hyper-parameter $\gamma$ in margin shifting (\ref{shifting}) are selected from $\{0, 0.01, 0.05\}$. All hyper-parameters are tuned on the development set, where those with the highest $F_1$ scores are selected for testing. All the experiments are conducted with 1 GeForce RTX 3090 GPU. We provide detailed configurations of the encoding models and the hyper-parameter settings on each dataset in Appendix. Our code is available at \url{https://github.com/yangzhou12/NCRL}.


\textbf{Competing methods.}
We compare our NCRL with existing multi-label loss functions including BCE and ATL \cite{zhou2021atlop}. We also test pairwise ranking loss and the log-sum-exp pairwise loss \cite{li2017improving}, but do not report their results for simplicity, as they are consistently worse than BCE in our experiments. For each method, we report the mean and standard deviation of performance measures based on 5 runs with different random seeds. The results with citations are taken from the corresponding papers. 

\subsection{Results and Discussions} \label{subsec:res}

\textbf{Results on DocRED.}
On DocRED, we follow the same setting of ATL and use its encoding model called ATLOP \cite{zhou2021atlop} for fair comparisons, which is built on RoBERTa-large~\cite{Liu2019RoBERTaAR}. To demonstrate our NCRL can still improve the performance on stronger encoding models, we also test a more recent pre-trained language model, DeBERTa-large \cite{he2021deberta}. In addition, we consider an improved RE model, which is first trained on the distantly supervised DocRED dataset and then finetuned on the human-annotated dataset. For comprehensiveness, we also compare recent document-level RE methods including HIN~\cite{Tang2020HINHI}, Coref~\cite{Ye2020CoreferentialRL}, CFER~\cite{dai2021coarse}, and SSAN~\cite{xu2021entity}, which use BCE for training. Following \cite{Yao2019DocREDAL}, we use $F_1$ and Ign $F_1$ in evaluation, where the Ign $F_1$ denotes the $F_1$ score excluding the relational facts that already appear in the training and dev/test sets. 

Table \ref{tab::docred_results} shows the document-level RE results on DocRED, where the test scores are obtained by submitting the predictions of the best checkpoint on the development set to the official Codalab evaluation system. Overall, BCE with a manually tuned threshold already achieves comparable or better results than most competing methods such as CFER and SSAN. By considering the none class in multi-label learning, ATL and our NCRL outperform BCE in most cases, and avoid the overhead of threshold tuning. This indicates that exploiting the correlations of the none class label can improve the performance of document-level RE. Moreover, our NCRL consistently outperforms ATL in all cases. This demonstrates the advantage of NCRL in learning a label ranking with maximized label margins. When the distantly supervised data are used for training, all the methods obtain much better results. In particular, NCRL achieves a state-of-the-art Ign $F_1$ score of $65.81\%$ on the leaderboard\footnote{\url{https://competitions.codalab.org/competitions/20717}}. This indicates that NCRL is robust against noisy labels and can better transfer the information from the distantly supervised dataset for document-level RE. On the other hand, ATL seems to be sensitive to label noise and gets much worse results than NCRL.

\begin{table*}[!t]
\centering
\small
    \begin{tabular}{lcccc}
         \toprule
         \textbf{Model} & \multicolumn{2}{c}{\textbf{Dev}} & \multicolumn{2}{c}{\textbf{Test}} \\
          & $F_1$ & ${F_1}_c$ & $F_1$ & ${F_1}_c$ \\
         \midrule
        NN~\cite{yu2020dialogue}        & 46.1 $\pm$ 0.7  & 43.7 $\pm$ 0.5    & 48.0 $\pm$ 1.5   & 45.0 $\pm$ 1.4  \\
        LSTM~\cite{yu2020dialogue}         & 46.7 $\pm$ 1.1  & 44.2 $\pm$ 0.8     & 47.4 $\pm$ 0.6  & 44.9 $\pm$ 0.7    \\
        BiLSTM~\cite{yu2020dialogue}      &  48.1 $\pm$ 1.0   & 44.3 $\pm$ 1.3     & 48.6 $\pm$ 1.0  & 45.0 $\pm$ 1.3    \\ 
        BERTs$_{\text{Base}}$~\cite{yu2020dialogue}  &  63.0 $\pm$1.5    & 57.3 $\pm$1.2    & 61.2 $\pm$0.9 & 55.4 $\pm$0.9 \\ 
        GDPNet~\cite{xue2021gdpnet}  &  67.1 $\pm$ 1.0    & 61.5 $\pm$ 0.8    & 64.9 $\pm$ 1.1 & 60.1 $\pm$ 0.9 \\ 
        \midrule
        BCE + BERTs$_{\text{Base}}$ &  66.10 $\pm$ 0.84    & 61.78 $\pm$ 0.68    & 63.25 $\pm$ 0.45 & 59.26 $\pm$ 0.37 \\ 
        ATL + BERTs$_{\text{Base}}$& 67.15 $\pm$ 0.75 & 62.66 $\pm$ 0.70 & \textbf{66.06 $\pm$ 1.09} & 61.62 $\pm$ 0.83 \\
        NCRL + BERTs$_{\text{Base}}$ & \textbf{68.10 $\pm$ 0.31} & \textbf{64.26 $\pm$ 0.37} & 66.00 $\pm$ 0.28 & \textbf{62.45 $\pm$ 0.22} \\
        \bottomrule
    \end{tabular}
    \caption{Classification results (\%) on the DialogRE dataset. }
    \label{tab::dialogre_results}
\end{table*}

\textbf{Results on DialogRE.}
On DialogRE, we adopt the official baseline, BERTs$_{\text{Base}}$ \cite{yu2020dialogue}, as the encoding model, where the dialogue inputs are augmented with entity type and coreference information. We follow \cite{yu2020dialogue} to use $F_1$ and ${F_1}_c$ in evaluation, where ${F_1}_c$ is the $F_1$ score computed by taking the first few turns instead of the entire dialogue as input. Table \ref{tab::dialogre_results} shows the dialogue-based RE results on DialogRE. As can be seen, our implementation of BERTs$_{\text{Base}}$ with BCE obtains significantly better results than the reported ones and is slightly worse than the state-of-the-art method, GDPNet, which combines the graph neural networks and BERTs$_{\text{Base}}$ for better representation learning and uses BCE for training. ATL and our NCRL outperform both BCE and GDPNet by a large margin, showing the benefit of learning the none class for document-level RE. Moreover, NCRL achieves the best and most stable results in most cases. This is because NCRL is more robust against imbalanced and mislabeled samples by applying the margin regularization and shifting approaches. It is worth noting that our margin regularization and shifting techniques are \textit{inapplicable} for BCE and ATL, as they do not formulate the probability of each class as label margins.

\textbf{Results on GoEmotions.}
For the GoEmotions dataset, sentences labeled with Neutral are considered as none class instances. Originally, there are a small number \textit{ambiguous} instances labeled with both Neutral and certain emotions. We remove these instances from the train/dev/test set for simplicity. We use BERT$_\text{Base}$ as the encoding model as in \cite{demszky-etal-2020-goemotions}, and finetune it with the plain NCRL (\ref{NCRL surrogate}) without any regularization, which is enough to achieve good performance in our experiments.
We use mean Average Precision (mAP) for evaluation, which summarizes a precision-recall curve at each threshold and measures the model ability in label ranking for each instance. Table \ref{tab::GoEmotions_results} shows the classification results on GoEmotions. 
Again, NCRL obtains the best results, indicating that NCRL works well on multi-label tasks when none class instances are available for training. Specifically, our NCRL improves BCE and ATL by $1.49\%$ and $1.25\%$ in mAP, respectively. 
This demonstrates that NCRL produces more accurate label rankings by capturing the correlations between the none class and pre-defined classes via the label margins.

\begin{table}[!t]
\centering
\small
    \begin{tabular}{lcc}
        \toprule
        \textbf{Model} & Dev mAP & Test mAP \\
        \midrule
        \emph{Encoding model} & \multicolumn{2}{c}{BERT$_{\text{Base}}$} \\
        BCE & 49.79 $\pm$ 0.62 & 48.28 $\pm$ 0.37 \\
        ATL & 49.95 $\pm$ 0.59 & 48.52 $\pm$ 0.78\\
        NCRL & \textbf{50.99 $\pm$ 0.35} & \textbf{49.77 $\pm$ 0.19}\\
        \bottomrule
    \end{tabular}
    \caption{Results (\%) on the GoEmotions dataset. }
    \label{tab::GoEmotions_results}
\end{table}

\textbf{Ablation studies.}
Finally, we perform ablation studies of NCRL on DialogRE. Table \ref{tab::ablation_results} provides the results of NCRL when each component of NCRL is excluded at a time, where M-Reg and M-shift denote the margin regularization and the margin shifting, respectively. In addition, we also compare BCE with probability shifting (P-shift) that is used in \cite{ben2020asymmetric,wu2020distribution} for multi-label learning. We can observe that all components of NCRL contribute to multi-label learning. Both the margin regularization and shifting techniques are important for NCRL to achieve good performance. Even without these two approaches, the plain NCRL still outperforms BCE and its variant, which implies that learning the none class from none class instances is indeed helpful for document-level RE.

\begin{table}[!t]
\centering
\small
    \begin{tabular}{lcccc}
        \toprule
        \textbf{Model} & \multicolumn{2}{c}{\textbf{Dev}} & \multicolumn{2}{c}{\textbf{Test}} \\
        & ${F_1}$ & ${F_1}_c$ & ${F_1}$ & ${F_1}_c$ \\
        \midrule
        \emph{NCRL} & 68.10 & 64.26 & 66.00 & 62.45 \\
        - M-Reg &  67.27  & 63.22 & 65.57 & 61.77 \\
        - M-Shift & 67.52 & 63.59 & 65.61 & 62.01 \\
        - Both & 67.56 & 63.04 & 65.86 & 61.65\\
        \emph{BCE} & 66.10 & 61.78 & 63.25 & 59.25 \\
        + P-Shift & 66.69 & 62.50 & 63.79 & 59.71 \\
        \bottomrule
    \end{tabular}
    \caption{Ablation studies on the DialogRE dataset. }
    \label{tab::ablation_results}
\end{table}

\section{Conclusion} \label{sec:conclusion}
We have proposed NCRL, a new multi-label loss that considers the probability of ``no relation" in document-level RE.
NCRL maximizes the label margin between the none class and each pre-defined class, capturing the label correlations of the none class and enabling adaptive thresholding for label prediction.
Through theoretical analysis, we justify the effectiveness of NCRL by proving that NCRL is Bayes consistent w.r.t. its targeted performance measure NCRE. 
In addition, a margin regularization and a margin shifting technique are proposed to gain robustness against imbalanced and mislabeled samples.
Experiments on three benchmark datasets demonstrate that NCRL significantly outperforms existing multi-label loss functions for document-level RE and emotion classification. 

\newpage
\appendix
\twocolumn

\section{Appendix}

\subsection{Proof of Bayes Consistency}
\textbf{Loss and risk.} In general, multi-label learning methods aim to find a classifier $\vech$ that minimizes the \textit{multi-label risk} $R_{\ell}(\vech) = \mathbb{E}_{(x,y)}[ \ell(\vecy, \vech(\vecx)) ]$, i.e., the expected classification loss over the joint distribution $\mathbb{P}(X, Y)$, where $\ell(\vecy, \vech(\vecx))$ is a \textit{multi-label loss} that measures the prediction performance of $\vech$. A multi-label classifier $\vech(\vecx)=(h_0(\vecx),\ldots,h_K(\vecx))$ is often induced from a \textit{score function} $\bm{f}(\vecx)=(f_0(\vecx),\ldots,f_K(\vecx))$ via \textit{thresholding}. Typically, one can take $h_i(\vecx) = \llbracket f_i(\vecx) > t(\vecx) \rrbracket$ with a thresholding function $t(\vecx)$, where $\llbracket \pi \rrbracket$ is the mapping that returns $1$ if predicate $\pi$ is true and $0$ otherwise. Classical multi-label performance measures include Hamming loss $\ell_H(\vecy,\vech)$ and ranking loss $\ell_{rnk}(\vecy,\vech)$, whose definitions are given as follows:
\begin{align*}
    \ell_H(\vecy,\vech)&=\sum_{i=1}^{K}  \llbracket y_i \neq h_i \rrbracket, \\
    \ell_{rnk}(\vecy,\vech)&=\sum_{(i,j): y_i > y_j} \Big( \llbracket f_i < f_j \rrbracket + \frac{1}{2} \llbracket f_i = f_j \rrbracket \Big).
\end{align*}
Intuitively, Hamming loss counts the number of misclassified labels, and ranking loss counts the number of reversely ordered label pairs.

\textbf{Surrogate loss and consistency.} Although our goal is to find an optimal classifier w.r.t. certain performance measure (loss), directly optimizing a performance measure such as Hamming loss and ranking loss is generally intractable. Because of this, one typically needs to minimize a \textit{surrogate loss} instead. For example, BCE and pairwise ranking loss are the surrogate losses of Hamming loss and ranking loss, respectively. A proper surrogate loss should be easy for optimization and \textit{consistent} with the corresponding performance measure. Formally, the multi-label consistency is defined as follows:

\begin{customdef}{1}[Multi-label consistency \cite{GAO201322} ]
The surrogate loss $L$ is Bayes consistent w.r.t. the multi-label performance measure $\ell$ if and only if it holds for any sequence $\bm{f}_n$ that
\begin{equation}
\label{eq:bayes-cls}
    R_{L}(\bm{f}_n) \rightarrow R^*_{\ell} \Rightarrow R_{\ell}(\bm{f}_n) \rightarrow \min R^*_{L},
\end{equation}
where $R^*_{\ell}$ and $R^*_{L}$ are the minimal risk w.r.t. $\ell$ and $L$, respectively.
\end{customdef}

Consistency is important for good classification performance, which guarantees that minimizing the surrogate loss can achieve the expected objective. Nevertheless, it is non-trivial to design a consistent surrogate loss. In fact, even some commonly used surrogate losses are inconsistent. For example, it is known that all convex surrogate loss functions including pairwise ranking loss is \textit{inconsistent} with the ranking loss \cite{GAO201322}. In what follows, we prove the Bayes consistency of NCRL w.r.t. NCRE.

\begin{customthm}{1}
    NCRL (\ref{NCRL surrogate}) is Bayes consistent w.r.t. NCRE (\ref{NCRL}).
\end{customthm}
\begin{proof}
    Let $\Delta_i = \mathbb{P}(y_i = 1|\vecx)$ be the marginal probability when the $i$-th label is positive. The conditional risk of NCRE (\ref{NCRL}) is given by
    \begin{align}
        R_{{L}_\textsc{na}}(\mathbb{P}, \bm{f}) = &\sum_{i=1}^{K} \Big( \Delta_i \cdot \log \sigma(f_i - f_0) \nonumber \\
        &+ (1 - \Delta_i) \cdot \log \sigma(f_0 - f_i) \Big).
    \end{align}
    For $i = 1, \ldots, K$, the partial derivative can be computed by
    \begin{align}
        \frac{\partial}{f_i} \mathbb{E}[ L_{\textsc{na}}(\mathbb{P}, \bm{f}) |\vecx] = &\Delta_i \frac{- \exp{(f_0 - f_i)}}{1 + \exp{(f_0 - f_i)}} \nonumber \\
        &+ (1 - \Delta_i) \frac{\exp{(f_i - f_0)}}{1 + \exp{(f_i - f_0)}}, 
    \end{align}
    Since NCRL (\ref{NCRL surrogate}) is convex and differentiable, we can obtain the optimal $\bm{f}^*$ by setting the partial derivatives to zero, which leads to
    \begin{equation}
        f_i^* - f_0^* = \log(\frac{\Delta_i}{1 - \Delta_i}), i = 1, \ldots, K.
    \end{equation}
    This means that, for the optimal score function $\bm{f}^*$, $f_i^* > f_0^*$ if and only if $\Delta_i > \frac{1}{2}$, which minimizes the NCRE risk according to (\ref{Bayes opt}). Therefore, NCRL is Bayes consistent w.r.t. NCRE (\ref{NCRL}).
\end{proof}

\subsection{Detailed Configurations of Encoding Models}
We construct the encoding model by finetuning a pre-trained language model (PLM) for representation learning. Since the inputs of different datasets are different, we follow existing works to build the encoding model for each dataset as follows:

\textbf{ATLOP on DocRED.}
We adopt ATLOP \cite{zhou2021atlop} as the encoding model on DocRED. Given a document $\mathcal{D}=[w_i]_{i=1}^L$ of length $L$, ATLOP inserts a special symbol ``*'' around each entity mentions in the document $\mathcal{D}$, and then feeds the document into a PLM to obtain $d$-dimensional contextual embeddings:
\begin{equation}
    \mathbf{H} = \left[\bm{h}_1, \bm{h}_2, ..., \bm{h}_L\right] = \text{PLM}(\left[w_1, w_2, ..., w_L\right]). \label{eq:encoder}
\end{equation}
In ATLOP, the embedding of ``*'' at the start of mentions is used as the mention embeddings.
For an entity $e_i$ with $N_{e_i}$ mentions $\{e_i^j\}_{j=1}^{N_{e_i}}$, the entity embedding $\bm{h}_{e_i}$ is computed by applying the logsumexp pooling~\cite{Jia2019DocumentLevelNR} as follows:
\begin{equation}
    \bm{h}_{e_i} = \log \sum_{j=1}^{N_{e_i}} \exp \left( \bm{h}_{e_i^j} \right). \label{eq:pool}
\end{equation}
Given the embeddings $(\bm{h}_{e_s}, \bm{h}_{e_o})$ of an entity pair $(e_s,e_o)$, the latent representations $\bm{z}_s$ and $\bm{z}_o$ of $e_s$ and $e_o$ are obtained by fusing $\bm{h}_{e_s}$ and $\bm{h}_{e_o}$ with a localized context embedding followed by a non-linear projection (please refer to \cite{zhou2021atlop} for more details). Finally, the entity pair representation is computed by $\bm{x} \in \mathbb{R}^{d^2} = \bm{z}_o \otimes \bm{z}_s$, and the confidence scores $\bm{f} \in \mathbb{R}^{K+1}$ given $\bm{x}$ is obtained by
\begin{equation}
    \bm{f} = \mathbf{W} \bm{x} + \bm{b}, 
\end{equation}
where $\otimes$ is the Kronecker product, $\bm{b} \in \mathbb{R}^{K+1}$ is a bias term, and $\mathbf{W} \in \mathbb{R}^{K+1 \times d^2}$ is the model parameters for the multi-label classifier.

\begin{table*}[!t]
\centering
\small
    \begin{tabular}{lcccc}
        \toprule
        \textbf{Model} & \multicolumn{2}{c}{\textbf{Dev}} & \multicolumn{2}{c}{\textbf{Test}} \\
        & $F_1$ & ${F_1}_c$ & $F_1$ & ${F_1}_c$ \\
        \midrule
        BCE &  66.10 $\pm$ 0.84    & 61.78 $\pm$ 0.68    & 63.25 $\pm$ 0.45 & 59.26 $\pm$ 0.37 \\ 
        BCE + Finer-Grained &  66.40 $\pm$ 0.55    & 62.03 $\pm$ 0.44   & 63.37 $\pm$ 0.42 & 59.36 $\pm$ 0.32 \\ 
        BCE + Per-Label &  67.65 $\pm$ 1.13    & 64.19 $\pm$ 0.58    & 62.91 $\pm$ 1.44 & 60.55 $\pm$ 0.91 \\ 
        ATL & 67.15 $\pm$ 0.75 & 62.66 $\pm$ 0.70 & \textbf{66.06 $\pm$ 1.09} & 61.62 $\pm$ 0.83 \\
        NCRL & \textbf{68.10 $\pm$ 0.31} & \textbf{64.26 $\pm$ 0.37} & 66.00 $\pm$ 0.28 & \textbf{62.45 $\pm$ 0.22} \\
        \bottomrule
    \end{tabular}
    \caption{Classification results (\%) on the DialogRE dataset. }
    \label{tab::dialogre_per_class_results}
\end{table*}

\textbf{BERTs on DialogRE.}
We adopt BERTs \cite{yu2020dialogue}, an improved BERT baseline, as the encoding model on DialogRE. Given a dialogue $\mathcal{D}$ and an entity pair $(e_s, e_o)$, BERTs augments the dialogue with the entity pair by [CLS] $\mathcal{D}$ [SEP] $e_s$ [SEP] $e_o$ [SEP], where [CLS] and [SEP] are the classification and separator tokens defined in BERT, respectively. If $e_s$ or $e_o$ indicates the name of a speaker, all the mentions of such speaker in the augmented dialogue are replaced by a special token [Sub] (or [Obj]) to locate the positions of relevant turns. After feeding the augmented dialogue into the BERT model, the embedding of [CLS] is used as the entity pair representation $\bm{x}$ for classification.

\textbf{BERT on GoEmotions.}
We follow \cite{demszky-etal-2020-goemotions} and use the plain BERT as the encoding model on GoEmotions for emotion classification. By feeding a given sentence into BERT, the embedding of [CLS] is used as the instance representation $\bm{x}$ for classification.

\subsection{Effect of Threshold Tuning}
In this section, we test BCE with carefully tuned thresholding and compare it with the proposed NCRL. Please note that NCRL and ATL can learn the threshold automatically, and only BCE needs to select the best threshold for prediction. Specifically, we consider two threshold tuning strategies for BCE: (1) We select a \textit{global} threshold from a finer-grained grid of $\{0.1, 0.11, ..., 0.9\}$, after confirming that the performance will not be further improved on the development set by searching more values. (2) We tune a \textit{per-label} threshold from $\{0.1, 0.11, ..., 0.9\}$.

Table \ref{tab::dialogre_per_class_results} shows the results of BCE with the threshold tuned by different strategies, where we also include the results reported in Table \ref{tab::dialogre_results} for clarity. As can be seen, with the finer-grained global threshold, BCE obtains slightly better results than the originally reported ones. When the pre-label threshold is used, BCE obtains much better results on the development set, whereas the improvement is not that significant on the test test. In particular, BCE even gets lower F1 scores than the originally reported results, indicating that the pre-label threshold overfits the development set. Even with a carefully tuned per-label threshold, BCE is still worse than our NCRL in all the cases. This demonstrates that by exploiting the none class instances, NCRL not only learns a better threshold but also achieves a better model 
for multi-label learning.

\begin{table}[!t]
\centering
\small
    \begin{tabular}{lcc}
        \toprule
        \textbf{Model} & Dev F1 & Test F1 \\
        \midrule
        BCE & 75.21 & \textbf{73.79} \\
        ATL & 73.93 & 72.25 \\
        ATL+ & 75.15 & 73.44 \\
        NCRL & 73.65 & 71.85 \\
        NCRL+ & \textbf{75.81} & 73.13 \\
        \bottomrule
    \end{tabular}
    \caption{Results (\%) on the AAPD dataset. }
    \label{tab::aapd}
\end{table}

\subsection{Importance of None Class Instances}
Although the proposed NCRL is a generic and can be used for multi-label tasks other than document-level RE, its effectiveness comes from the information contained in none class instances. 
When there is no none class instance for training, it is difficult for NCRL to estimate the confidence score of the none class and capture correct correlations between the none class and pre-defined classes via label ranking.
Without positive samples of $y_0=1$, the score $f_0$ does not reflect the confidence for an instanced to be labeled as the none class anymore, as $f_0$ will always be ranked below certain pre-defined labels during training. 
Consequently, NCRL losses its ability in adjusting label rankings and performing adaptive thresholding.

To study how none class instances affect the performance of NCRL, we conduct experiments on the AAPD dataset \cite{yang-etal-2018-sgm} for multi-label text classification, which contains the abstract of 55840 papers from 54 subjects in the field of computer science. All the abstracts in the AAPD dataset have at least one subject (pre-defined label), and there is no none class instance. We use the BERT base pre-trained model for text encoding and then finetune it for classification. 

Table \ref{tab::aapd} shows the classification results on AAPD. ATL+ and NCRL+ represent ATL and NCRL using a tuned threshold as in BCE rather than the none class label for multi-label prediction, respectively. As can be seen, when the learned threshold is used, ATL and NCRL are worse than BCE. On the other hand, when the tuned threshold is used, all the losses obtain similar results. This implies that the lack of none class instances prevents ATL and NCRL from learning the correct confidence scores of the none class for thresholding. Different from AAPD, document-level relation extraction datasets contain many none class instances, which can be utilized by NCRL for better classification performance.

\subsection{Sensitivity to Margin Shifting}
\begin{figure}[t!]
\centering
\includegraphics[width=0.8\columnwidth]{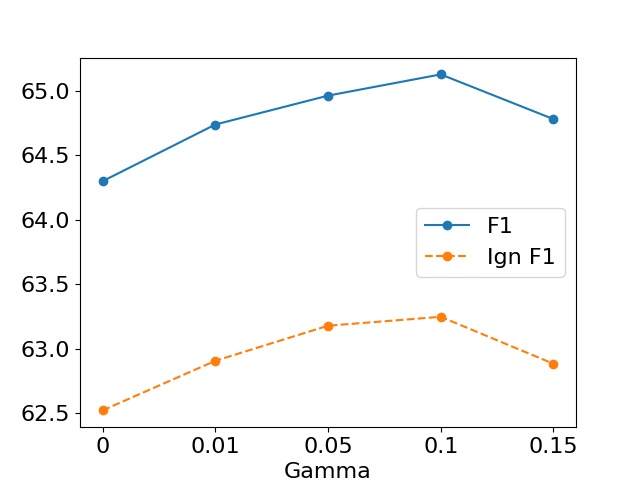} 
\caption{Classification results of NCRL with different $\gamma$ on DocRED.}
\label{F1vsGamma}
\end{figure}

Figure \ref{F1vsGamma} shows the performance of NCRL with different $\gamma$ on the DocRED dataset. As can be seen, NCRL is not very sensitive to $\gamma$.

\subsection{Hyper-Parameter Settings}
Table \ref{tab::hyper} provides the detailed hyper-parameter settings on different datasets.

\begin{table*}[!t]
\centering
\small
    \begin{tabular}{lcccccc}
         \toprule
         \textbf{Hyper-param}& \multicolumn{3}{c}{\textbf{DocRED}} & \multicolumn{1}{c}{\textbf{DialogRE}}& \multicolumn{1}{c}{\textbf{GoEmotions}} & \multicolumn{1}{c}{\textbf{AAPD}}\\
         & BERT& RoBERTa & DeBERTa & BERT & BERT & BERT\\
         \midrule
         Batch size& 4 & 4 & 4 & 24 & 16 & 16\\
         Gradient accumulation steps & 1 & 2 & 2 & 6 & 1 & 1 \\
         \# Epoch& 8 & 8 & 8 & 20 & 10 & 30 \\
         lr for encoder& 5e-5& 3e-5& 1e-5& 3e-5 & 5e-5 & 2e-5  \\
         lr for classifier& 1e-4& 1e-4& 1e-4& 3e-5 & 5e-5 & 2e-5 \\
         $\gamma$ for margin shifting  & 0.05 & 0.05& 0.05& 0.01 & 0 & 0.01 \\
         \bottomrule
    \end{tabular}
    \caption{Hyper-parameter settings on different datasets.}  \label{tab::hyper}
\end{table*}

\section*{Acknowledgments}
This research is supported in part by the National Research Foundation, Singapore under its AI
Singapore Program (AISG Award No: AISG2-RP-2020-016).

\small
\bibliographystyle{named}
\bibliography{ijcai22}

\end{document}